# Efficient Clustering with Limited Distance Information

**Konstantin Voevodski**
Dept. of Computer Science
Boston University
Boston, MA 02215

**Maria-Florina Balcan**
College of Computing
Georgia Institute of Technology
Atlanta, GA 30332

**Heiko Röglin**
Dept. of Quantitative Economics
Maastricht University
Maastricht, The Netherlands

**Shang-Hua Teng**
Computer Science Dept.
University of Southern California
Los Angeles, CA 90089

**Yu Xia**
Bioinformatics Program and Dept. of Chemistry
Boston University
Boston, MA 02215

## Abstract

Given a point set $S$ and an unknown metric $d$ on $S$, we study the problem of efficiently partitioning $S$ into $k$ clusters while querying few distances between the points. In our model we assume that we have access to *one versus all* queries that given a point $s \in S$ return the distances between $s$ and all other points. We show that given a natural assumption about the structure of the instance, we can efficiently find an accurate clustering using only $O(k)$ distance queries. We use our algorithm to cluster proteins by sequence similarity. This setting nicely fits our model because we can use a fast sequence database search program to query a sequence against an entire dataset. We conduct an empirical study that shows that even though we query a small fraction of the distances between the points, we produce clusterings that are close to a desired clustering given by manual classification.

## 1 Introduction

Clustering from pairwise distance information is an important problem in the analysis and exploration of data. It has many variants and formulations and it has been extensively studied in many different communities, and many different clustering algorithms have been proposed.

Many application domains ranging from computer vision to biology have recently faced an explosion of data, presenting several challenges to traditional clustering techniques. In particular, computing the distances between all pairs of points, as required by traditional clustering algorithms, has become infeasible in many application domains. As a consequence it has become increasingly important to develop effective clustering algorithms that can operate with limited distance information.

In this work we initiate a study of clustering with limited distance information; in particular we consider clustering with a small number of *one versus all* queries. We can imagine at least two different ways to query distances between points. One way is to ask for distances between pairs of points, and the other is to ask for distances between one point and all other points. Clearly, a one versus all query can be implemented as $|S|$ pairwise queries, but we draw a distinction between the two because the former is often significantly faster in practice if the query is implemented as a database search.

Our main motivating example for considering one versus all distance queries is sequence similarity search in biology. A program such as BLAST [1] (Basic Local Alignment Search Tool) is optimized to search a single sequence against an entire database of sequences. On the other hand, performing $|S|$ pairwise sequence alignments takes several orders of magnitude more time, even if the pairwise alignment is very fast. The disparity in runtime is due to the hashing that BLAST uses to identify regions of similarity between the input sequence and sequences in the database. The program maintains a hash table of all *words* in the database (substrings of a certain length), linking each word to its locations. When a query is performed, BLAST considers each word in the input sequence, and runs a local sequence alignment in each of its locations in the database. Therefore the program only performs a limited number of local sequence alignments, rather than aligning the input sequence to each sequence in the database. Of course, the downside is that we never consider alignments between sequences that do not share a word. However, in this case an alignment may not be relevant anyway, and we can assign a distance of infinity to the two sequences. Even though the search performed by BLAST is heuristic, it has been shown that protein sequence similarity identified by BLAST is meaningful [4].

Motivated by such scenarios, in this paper we consider the problem of clustering a dataset with an unknown distance function, given only the capability to ask one versus all distance queries. We design an efficient algorithm for clustering accurately with a small number of such queries. To formally analyze the correctness of our algorithm we assume that the distance function is a metric, and that our clustering problem satisfies a natural approximation stability property regarding the utility of the $k$-median objective function in clustering the points. In particular, our analysis assumes the $(c, \epsilon)$-property of Balcan et al. [3]. For an objective function $\Phi$ (such as k-median), the $(c, \epsilon)$-property assumes that any clustering that is a $c$-approximation of $\Phi$ has error of at most $\epsilon$. To define what we mean by error we assume that there exists some unknown relevant "target" clustering $C_T$; the error of a proposed clustering $C$ is then the fraction of misclassified points under the optimal matching between the clusters in $C_T$ and $C$.

Our first main contribution is designing an algorithm that given the $(c, \epsilon)$-property for the k-median objective finds a clustering that is very close to the target by using only $O(k)$ one versus all queries. In particular, we use the same assumption as Balcan et al. [3], and we obtain the same performance guarantees as [3] but by only using a very small number of one versus all queries. In addition to handling this more difficult scenario, we also provide a much faster algorithm. The algorithm of [3] can be implemented in $O(|S|^3)$ time, while the one proposed here runs in time $O(|S|k \log |S|)$.

We also use our algorithm to cluster proteins by sequence similarity, and compare our results to gold standard manual classifications given in the Pfam [6] and SCOP [8] databases. These classification databases are used ubiquitously in biology to observe evolutionary relationships between proteins and to find close relatives of particular proteins. We find that for one of these sources we obtain clusterings that usually closely match the given classification, and for the other the performance of our algorithm is comparable to that of the best known algorithms using the full distance matrix. Both of these classification databases have limited coverage, so a completely automated method such as ours can be useful in clustering proteins that have yet to be classified. Moreover, our method can cluster very large datasets because it is efficient and does not require the full distance matrix as input, which may be infeasible to obtain for a very large dataset.

**Related Work:** A property that is related to $(c, \epsilon)$ is $\epsilon$-separability, introduced by Ostrovsky et al. [9]. A clustering instance is $\epsilon$-separated if the cost of the optimal $k$-clustering is at most $\epsilon^2$ times the cost of the optimal clustering using $k-1$ clusters. The $\epsilon$-separability and $(c, \epsilon)$ properties are related: in the case when the clusters are large the Ostrovsky et al. [9] condition implies the Balcan et al. [3] condition (see [3]).

Ostrovsky et al. also present a sampling method for choosing initial centers, which when followed by a single Lloyd-type descent step gives a constant factor approximation of the $k$-means objective if the instance is $\epsilon$-separated. However, their sampling method needs information about the full distance matrix because the probability of picking two points as two cluster centers is proportional to their squared distance. A very similar (independently proposed) strategy is used in [2] to obtain an $O(\log k)$-approximation of the $k$-means objective on arbitrary instances. This strategy can be implemented with $k$ one versus all distance queries. However, an $O(\log k)$-approximation is not good enough for our purposes.

Approximate clustering using sampling has been studied extensively in recent years (see, e.g., [7, 5]). The methods proposed in these papers yield constant factor approximations to the $k$-median objective using $O(k)$ one versus all distance queries. However, as the constant factor of these approximations is at least 2, the proposed sampling methods do not necessarily yield clusterings close to the target clustering $C_T$ if the $(c, \epsilon)$-property holds only for some small constant $c < 2$, which is the interesting case in our setting.

## 2 Preliminaries

Given a metric space $M = (X, d)$ with point set $X$, an unknown distance function $d$ satisfying the triangle inequality, and a set of points $S \subseteq X$, we would like to find a $k$-clustering $C$ that partitions the points in $S$ into $k$ sets $C_1, C_2, \ldots C_k$ by using *one versus all* distance queries.

In our analysis we assume that $S$ satisfies the $(c, \epsilon)$-property of Balcan et al. [3] for the $k$-median objective function. The $k$-median objective is to minimize $\Phi(C) = \sum_{i=1}^{k} \sum_{x \in C_i} d(x, c_i)$, where $c_i$ is the median of cluster $C_i$, which is the point $y$ that minimizes $\sum_{x \in C_i} d(x, y)$. Let $\mathrm{OPT}_\Phi = \min_C \Phi(C)$, where the minimum is over all $k$-clusterings of $S$, and denote by $C^* = \{C_1^*, C_2^*, \ldots C_k^*\}$ a clustering achieving this value.

To formalize the $(c, \epsilon)$-property we need to define a notion of distance between two $k$-clusterings $C = \{C_1, C_2, \ldots C_k\}$ and $C' = \{C_1', C_2', \ldots C_k'\}$. As in [3], we define the distance between $C$ and $C'$ as the fraction of points on which they disagree under the optimal matching of clusters in $C$ to clusters in $C'$: $\text{dist}(C, C') = \min_{\sigma \in S_k} \frac{1}{n} \sum_{i=1}^{k} |C_i - C'_{\sigma(i)}|$, where $S_k$ is the set of bijections $\sigma \colon [k] \to [k]$. Two clusterings $C$ and $C'$ are $\epsilon$-close if $\text{dist}(C, C') < \epsilon$.

We assume that there exists some unknown relevant "target" clustering $C_T$ and given a proposed clustering $C$ we define the error of $C$ with respect to $C_T$ as $\text{dist}(C, C_T)$. Our goal is to find a clustering of low error.

The $(c, \epsilon)$-property is defined as follows.

**Definition 1.** *We say that the instance $(S, d)$ satisfies the $(c, \epsilon)$-property for the $k$-median objective function with respect to the target clustering $C_T$ if any clustering of $S$ that approximates $\text{OPT}_\Phi$ within a factor of $c$ is $\epsilon$-close to $C_T$, i.e., $\Phi(C) \leq c \cdot \text{OPT}_\Phi \Rightarrow \text{dist}(C, C_T) < \epsilon$.*

In the analysis of the next section we denote by $c_i^*$ the center point of $C_i^*$, and use OPT to refer to the value of $C^*$ using the $k$-median objective, i.e., $\text{OPT} = \Phi(C^*)$. We define the *weight* of point $x$ to be the contribution of $x$ to the $k$-median objective in $C^*$: $w(x) = \min_i d(x, c_i^*)$. Similarly, we use $w_2(x)$ to denote $x$'s distance to the second-closest cluster center among $\{c_1^*, c_2^*, \ldots, c_k^*\}$. In addition, let $w$ be the average weight of the points: $w = \frac{1}{n} \sum_{i=1}^{n} w(x) = \frac{\text{OPT}}{n}$, where $n$ is the cardinality of $S$.

## 3 Clustering With Limited Distance Information

---
**Algorithm 1** Landmark-Clustering$(S, d, \alpha, \epsilon, k)$

$b = (1 + 17/\alpha)\epsilon n$;
$q = 2b$;
$\text{iter} = 4k$;
$s_{\min} = b + 1$;
$n' = n - b$;
$L = \textbf{Landmark-Selection}(q, \text{iter})$;
$C' = \textbf{Expand-Landmarks}\ (s_{\min}, n', L)$;
Choose some landmark $l_i$ from each cluster $C_i'$;
**for** each $x \in S$ **do**
    Insert $x$ into the cluster $C_j''$ for $j = \text{argmin}_i d(x, l_i)$;
**end for**
**return** $C''$;

---

In this section we present a new algorithm that accurately clusters a set of points assuming that the clustering instance satisfies the $(c, \epsilon)$-property for $c = 1 + \alpha$, and the clusters in the target clustering $C_T$ are not too small. The algorithm presented here is much faster than the one given by Balcan et al., and does not require all pairwise distances as input. Instead, we only require $O(k)$ one versus all distance queries to achieve the same performance guarantee as in [3].

Our clustering algorithm is described in Algorithm 1. We start by using the *Landmark-Selection* procedure to select a set of $4k$ landmarks. This procedure repeatedly chooses uniformly at random one of the $q$ furthest points from the ones selected so far, for an appropriate $q$. Our algorithm only uses the distances between the selected landmarks and other points, so it requires only $4k$ one versus all distance queries.

---
**Algorithm 2** Landmark-Selection$(q, \text{iter})$

Choose $s^* \in S$ uniformly at random;
$L = \{s^*\}$;
**for** $i = 1$ to $\text{iter} - 1$ **do**
    Let $s_1, \ldots, s_n$ be an ordering of the points in $S$ such that $\min_{l \in L} d(l, s_i) \leq \min_{l \in L} d(l, s_{i+1})$;
    Choose $s^* \in \{s_{n-q+1}, \ldots, s_n\}$ uniformly at random;
    $L = L \cup \{s^*\}$;
**end for**
**return** $L$;

---

It is possible to implement each iteration of *Landmark-Selection* in $O(n)$ time: For each point we store the minimum distance to the landmarks chosen so far, which is updated in constant time when we add a new landmark. To select a new landmark in each iteration, we choose a random number $i \in [n - q + 1, n]$ and use a linear time selection algorithm to select the $i$th furthest point as the next landmark.

*Expand-Landmarks* then expands a ball $B_l$ around each landmark $l \in L$ chosen by *Landmark-Selection*. We use the variable $r$ to denote the radius of all the balls: $B_l = \{s \in S \mid d(s, l) \leq r\}$. The algorithm starts with $r = 0$, and increments it until the balls satisfy a property described below. For each $B_l$ there are $n$ relevant values of $r$ to try, each adding one more point to $B_l$, which results in at most $|L|n$ values to try in total.

The algorithm maintains a graph $G_B = (V_B, E_B)$, where vertices correspond to balls that have at least $s_{\min}$ points in them, and two vertices are connected by an (undirected) edge if the corresponding balls overlap on any point: $(v_{l_1}, v_{l_2}) \in E_B$ iff $B_{l_1} \cap B_{l_2} \neq \emptyset$. In addition, we maintain the set of points in these balls

Clustered $= \{s \in S \mid \exists l\colon s \in B_l\}$ and a list of the connected components of $G_B$, which we refer to as Components$(G_B) = \{\text{Comp}_1, ..., \text{Comp}_m\}$.

In each iteration, after we expand one of the balls by a single point, we update $G_B$, Components$(G_B)$, and Clustered. If $G_B$ has exactly $k$ components, and $|\text{Clustered}| \geq n'$, we terminate and report points in balls that are part of the same component in $G_B$ as distinct clusters. If this condition is never satisfied, we report **no-cluster**. A sketch of the algorithm is given below. We use $(l^*, s^*)$ to refer to the next landmark-point pair that is considered, corresponding to expanding $B_{l^*}$ to include $s^*$.

---
**Algorithm 3** Expand-Landmarks$(s_{\min}, n', L)$
---
1: **while** $((l^*, s^*) = $ Expand-Ball$()) \mathrel{!=} $ null **do**
2:     $r = d(l^*, s^*)$;
3:     update $G_B$, Components$(G_B)$, and Clustered
4:     **if** $|\text{Components}(G_B)| = k$ and $|\text{Clustered}| \geq n'$ **then**
5:        **return** $C = \{C_1, ..., C_k\}$ where $C_i = \{s \in S \mid \exists l\colon s \in B_l \text{ and } v_l \in \text{Comp}_i\}$.
6:     **end if**
7: **end while**
8: **return no-cluster**;

---

Using a min-heap to store all landmark-point pairs and a disjoint-set data structure to keep track of the connected components of $G_B$, each iteration of the while loop can be implemented in amortized time $O(\log n)$. As the number of iterations is bounded by $|L|n$, this gives a worst-case running time of $O(|L|n \log n)$.[1]

The last step of our algorithm takes the clustering $C'$ returned by *Expand-Landmarks* and improves it: We compute a set $L'$ that contains exactly one landmark from each cluster $C'_i \in C'$ (any landmark is sufficient), and assign each point $x \in S$ to the cluster corresponding to the closest landmark in $L'$.

The runtime of *Landmark-Selection* is $O(kn)$, *Expand-Landmarks* can be implemented in $O(kn \log n)$, and the last part of the procedure takes $O(kn)$ time, thus the total runtime of the algorithm is $O(kn \log n)$. Moreover, the algorithm only uses the distances between the selected landmarks and other points, so it only uses $O(k)$ one versus all distance queries.

We now present our main theoretical guarantee for Algorithm 1.

**Theorem 2.** *Given a metric space $M = (X, d)$, where $d$ is unknown, and a set of points $S$, if the instance $(S, d)$ satisfies the $(1 + \alpha, \epsilon)$-property for the k-median objective function and if each cluster in the target clustering $C_T$ has size at least $(4+51/\alpha)\epsilon n$, then with probability $1 - \exp(-\Omega(k))$ Landmark-Clustering outputs a clustering that is $\epsilon$-close to $C_T$.*

Before we prove the theorem, we will introduce some notation and use an analysis similar to the one in [3] to argue about the structure of the clustering instance. Let $\epsilon^* = \text{dist}(C_T, C^*)$. By our assumption that the k-median clustering of $S$ satisfies the $(1 + \alpha, \epsilon)$-property we have $\epsilon^* < \epsilon$. Since each cluster in the target clustering has at least $(4+51/\alpha)\epsilon n$ points, and the *optimal k-median clustering* $C^*$ differs from the target clustering by $\epsilon^* n \leq \epsilon n$ points, each cluster in $C^*$ must have at least $(3 + 51/\alpha)\epsilon n$ points.

Let us define the *critical distance* $d_{\text{crit}} = \frac{\alpha w}{17\epsilon}$. We call a point $x$ *good* if both $w(x) < d_{\text{crit}}$ and $w_2(x) - w(x) \geq 17 d_{\text{crit}}$, else $x$ is called *bad*. In other words, the *good* points are those points that are close to their own cluster center and far from any other cluster center. In addition, we will break up the *good* points into *good sets* $X_i$, where $X_i$ is the set of the *good* points in the optimal cluster $C_i^*$. So each set $X_i$ is the "core" of the optimal cluster $C_i^*$.

Note that the distance between two points $x, y \in X_i$ satisfies $d(x, y) \leq d(x, c_i^*) + d(c_i^*, y) = w(x) + w(y) < 2 d_{\text{crit}}$. In addition, the distance between any two points in different good sets is greater than $16 d_{\text{crit}}$. To see this, consider a pair of points $x \in X_i$ and $y \in X_{j \neq i}$. The distance from $x$ to $y$'s cluster center $c_j^*$ is at least $17 d_{\text{crit}}$. By the triangle inequality, $d(x, y) \geq d(x, c_j^*) - d(y, c_j^*) > 17 d_{\text{crit}} - d_{\text{crit}} = 16 d_{\text{crit}}$.

It is proved in [3] that if the k-median instance $(M, S)$ satisfies the $(1+\alpha, \epsilon)$-property with respect to $C_T$, and each cluster in $C_T$ has size at least $2\epsilon n$, then

1. less than $(\epsilon - \epsilon^*)n$ points $x \in S$ on which $C_T$ and $C^*$ agree have $w_2(x) - w(x) < \frac{\alpha w}{\epsilon}$.

2. at most $17\epsilon n/\alpha$ points $x \in S$ have $w(x) \geq \frac{\alpha w}{17\epsilon}$.

The intuition is that if too many points on which $C_T$ and $C^*$ agree are close enough to the second-closest center among $\{c_1^*, c_2^*, \ldots, c_k^*\}$, then we can move them to the clusters corresponding to those centers, producing a clustering that is far from $C_T$, but whose objective value is close to OPT, violating the $(1 + \alpha, \epsilon)$-property. The second part follows from the fact that $\sum_{x \in S} w(x) = OPT = wn$.

Then using these facts and the definition of $\epsilon^*$ it follows that at most $\epsilon^* n + (\epsilon - \epsilon^*)n + 17\epsilon n/\alpha = \epsilon n + 17\epsilon n/\alpha = (1 + 17/\alpha)\epsilon n = b$ points are bad. Hence each $|X_i| = |C_i^* \setminus B| \geq (2 + 34/\alpha)\epsilon n = 2b$.

---
[1] A detailed description of this implementation is given in the full version of this paper, which can be downloaded from http://cs-people.bu.edu/kvodski/UAI10.pdf.

In the remainder of this section we prove that given this structure of the clustering instance, *Landmark-Clustering* finds an accurate clustering. We first show that almost surely the set of landmarks returned by *Landmark-Selection* has the property that each of the cluster cores has a landmark near it. We then argue that *Expand-Landmarks* finds a partition $C'$ that clusters most of the points in each core correctly. We conclude with the proof of the theorem, which argues that the clustering returned by the last step of our procedure is a further improved clustering that is very close to $C^*$ and $C_T$.

The *Landmark-Clustering* algorithm first uses *Landmark-Selection*$(q, iter)$ to choose a set of landmark points. The following lemma proves that if $q = 2b$ and $iter = 4k$ almost surely the set of selected landmarks has the property that there is a landmark closer than $2d_{\text{crit}}$ to some point in each good set.

**Lemma 3.** *Given $L$ = Landmark-Selection $(2b, 4k)$, with probability $1 - \exp(-\Omega(k))$ there is a landmark closer than $2d_{crit}$ to some point in each good set.*

*Proof.* Because there are at most $b$ bad points and in each iteration we uniformly at random choose one of $2b$ points, the probability that a good point is added to $L$ is at least $1/2$ in each iteration. Using a Chernoff bound we show that the probability that fewer than $k$ good points have been added to $L$ after $t$ iterations is less than $e^{-t(1-\frac{2k}{t})^2/4}$ (Lemma 4). Therefore after $4k$ iterations $k$ good points have been added to $L$ with probability $1 - e^{-\Omega(k)}$. Note that these good points must be distinct because we cannot choose the same point twice in the first $n - 2b$ iterations. There are two possibilities regarding the first $k$ good points added to $L$: they are either selected from distinct good sets, or at least two of them are selected from the same good set.

If the former is true then the statement trivially holds. If the latter is true, consider the first time that a second point is chosen from the same good set $X_i$. Let us call these two points $x$ and $y$, and assume that $y$ is chosen after $x$. The distance between $x$ and $y$ must be less than $2d_{\text{crit}}$ because they are in the same good set. Therefore when $y$ is chosen, $\min_{l \in L} d(l, y) \leq d(x, y) < 2d_{\text{crit}}$. Moreover, $y$ is chosen from $\{s_{n-2b+1}, ..., s_n\}$, where $\min_{l \in L} d(l, s_i) \leq \min_{l \in L} d(l, s_{i+1})$. Therefore when $y$ is chosen, at least $n - 2b + 1$ points $s \in S$ (including $y$) satisfy $\min_{l \in L} d(l, s) \leq \min_{l \in L} d(l, y) < 2d_{\text{crit}}$. Since each good set satisfies $|X_i| \geq 2b$, it follows that there must be a landmark closer than $2d_{crit}$ to some point in each good set. □

**Lemma 4.** *The probability that fewer than $k$ good points have been chosen as landmarks after $t \geq 2k$ iterations of Landmark-Selection is less than $e^{-t(1-\frac{2k}{t})^2/4}$.*

*Proof.* Let $X_i$ be an indicator random variable defined as follows: $X_i = 1$ if point chosen in iteration $i$ is a good point, and 0 otherwise. Let $X = \sum_{i=1}^{t} X_i$, and $\mu$ be the expectation of $X$. In other words, $X$ is the number of good points chosen after $t$ iterations of the algorithm, and $u$ is its expected value.

Because in each round we uniformly at random choose one of $2b$ points and there are at most $b$ bad points in total, $\mathrm{E}[X_i] \geq 1/2$ and hence $\mu \geq t/2$. By the Chernoff bound, for any $\delta > 0$ $\Pr[X < (1-\delta)\mu] < e^{-\mu\delta^2/2}$.

If we set $\delta = 1 - \frac{2k}{t}$, we have $(1-\delta)\mu = (1-(1-\frac{2k}{t}))\mu \geq (1-(1-\frac{2k}{t}))t/2 = k$. Assuming that $t \geq 2k$, it follows that $\Pr[X < k] \leq \Pr[X < (1-\delta)\mu] < e^{-\mu\delta^2/2} = e^{-\mu(1-\frac{2k}{t})^2/2} \leq e^{-t/2(1-\frac{2k}{t})^2/2}$. □

The algorithm then uses the *Expand-Landmarks* procedure to find a $k$-clustering $C'$. The following lemma states that $C'$ is an accurate clustering, and has an additional property that is relevant for the last part of the algorithm.

**Lemma 5.** *Given a set of landmarks $L$ chosen by Landmark-Selection so that the condition in Lemma 3 is satisfied, Expand-Landmarks$(b+1, n-b, L)$ returns a $k$-clustering $C' = \{C'_1, C'_2, \ldots C'_k\}$ in which each cluster contains points from a distinct good set $X_i$. If we let $\sigma$ be a bijection mapping each good set $X_i$ to the cluster $C'_{\sigma(i)}$ containing points from $X_i$, the distance between $c^*_i$ and any landmark $l$ in $C'_{\sigma(i)}$ satisfies $d(c^*_i, l) < 5d_{\text{crit}}$.*

*Proof.* Lemma 6 argues that since the good sets $X_i$ are well-separated, for $r < 4d_{\text{crit}}$ no ball of radius $r$ can overlap more than one $X_i$, and two balls that overlap different $X_i$ cannot share any points. Moreover, since we only consider balls that have more than $b$ points in them, and the number of bad points is at most $b$, each ball in $G_B$ must overlap some good set. Lemma 7 argues that since there is a landmark near each good set, there is a value of $r^* < 4d_{\text{crit}}$ such that each $X_i$ is contained in some ball around a landmark of radius $r^*$. We can use these facts to argue for the correctness of the algorithm.

First we observe that for $r = r^*$, $G_B$ has exactly $k$ components and each good set $X_i$ is contained within a distinct component. Each ball in $G_B$ overlaps with some $X_i$, and by Lemma 6, since $r^* < 4d_{\text{crit}}$, we know that each ball in $G_B$ overlaps with exactly one $X_i$. From Lemma 6 we also know that balls that overlap different $X_i$ cannot share any points and are thus not connected in $G_B$. Therefore balls that overlap different

$X_i$ will be in different components in $G_B$. Moreover, by Lemma 7 each $X_i$ is contained in some ball of radius $r^*$. For each good set $X_i$ let us designate by $B_i$ a ball that contains all the points in $X_i$, which is in $G_B$ since the size of each good set satisfies $|X_i| > b$. Any ball in $G_B$ that overlaps $X_i$ will be connected to $B_i$, and will thus be in the same component as $B_i$. Therefore for $r = r^*$, $G_B$ has exactly $k$ components, one for each good set $X_i$ that contains all the points in $X_i$.

Since there are at least $n - b$ good points that are in some $X_i$, this means that for $r = r^*$ the number of points that are in some ball in $G_B$ (which are in Clustered) is at least $n - b$. Hence the condition in line 4 of *Expand-Landmarks* will be satisfied and the algorithm will terminate and return a $k$-clustering in which each cluster contains points from a distinct good set $X_i$.

Now let us suppose that we start with $r = 0$. Consider the first value of $r = r'$ for which the condition in line 4 is satisfied. At this point $G_B$ has exactly $k$ components and the number of points that are not in these components is at most $b$. It must be the case that $r' \leq r^* < 4d_{\text{crit}}$ because we know that the condition is satisfied for $r = r^*$, and we are considering all relevant values of $r$ in ascending order. As before, each ball in $G_B$ must overlap some good set $X_i$. Again using Lemma 6 we argue that since $r < 4d_{\text{crit}}$, no ball can overlap more than one $X_i$ and two balls that overlap different $X_i$ cannot share any points. It follows that each component of $G_B$ contains points from a single $X_i$ (so we cannot merge the good sets). Moreover, since the size of each good set satisfies $|X_i| > b$, and there are at most $b$ points left out of $G_B$, each component must contain points from a distinct $X_i$ (so we cannot split the good sets). Thus we will return a $k$-clustering in which each cluster contains points from a distinct good set $X_i$.

To prove the second part of the statement, let $\sigma$ be a bijection matching each good set $X_i$ to the cluster $C'_{\sigma(i)}$ containing points from $X_i$. Clearly, $C'_{\sigma(i)}$ is made up of points in balls of radius $r < 4d_{\text{crit}}$ that overlap $X_i$. Consider any such ball $B_l$ around landmark $l$ and let $s^*$ denote any point on which $B_l$ and $X_i$ overlap. By the triangle inequality, the distance between $c_i^*$ and $l$ satisfies $d(c_i^*, l) \leq d(c_i^*, s^*) + d(s^*, l) < d_{\text{crit}} + r < 5d_{\text{crit}}$. Therefore the distance between $c_i^*$ and any landmark $l \in C'_{\sigma(i)}$ satisfies $d(c_i^*, l) < 5d_{\text{crit}}$. $\square$

**Lemma 6.** *A ball of radius $r < 4d_{\text{crit}}$ cannot contain points from more than one good set $X_i$, and two balls of radius $r < 4d_{\text{crit}}$ that overlap different $X_i$ cannot share any points.*

*Proof.* To prove the first part, consider a ball $B_l$ of radius $r < 4d_{\text{crit}}$ around landmark $l$. In other words, $B_l = \{s \in S \mid d(s, l) \leq r\}$. If $B_l$ overlaps more than one good set, then it must have at least two points from different good sets $x \in X_i$ and $y \in X_j$. By the triangle inequality it follows that $d(x, y) \leq d(x, l) + d(l, y) \leq 2r < 8d_{\text{crit}}$. However, we know that $d(x, y) > 16d_{\text{crit}}$, giving a contradiction.

To prove the second part, consider two balls $B_{l_1}$ and $B_{l_2}$ of radius $r < 4d_{\text{crit}}$ around landmarks $l_1$ and $l_2$. In other words, $B_{l_1} = \{s \in S \mid d(s, l_1) \leq r\}$, and $B_{l_2} = \{s \in S \mid d(s, l_2) \leq r\}$. Assume that they overlap with different good sets $X_i$ and $X_j$: $B_{l_1} \cap X_i \neq \emptyset$ and $B_{l_2} \cap X_j \neq \emptyset$. For the purpose of contradiction, let's assume that $B_{l_1}$ and $B_{l_2}$ share at least one point: $B_{l_1} \cap B_{l_2} \neq \emptyset$, and use $s^*$ to refer to this point. By the triangle inequality, it follows that the distance between any point $x \in B_{l_1}$ and $y \in B_{l_2}$ satisfies $d(x, y) \leq d(x, s^*) + d(s^*, y) \leq [d(x, l_1) + d(l_1, s^*)] + [d(s^*, l_2) + d(l_2, y)] \leq 4r < 16d_{\text{crit}}$.

Since $B_{l_1}$ overlaps with $X_i$ and $B_{l_2}$ overlaps with $X_j$, it follows that there is a pair of points $x \in X_i$ and $y \in X_j$ such that $d(x, y) < 16d_{\text{crit}}$, a contradiction. Therefore if $B_{l_1}$ and $B_{l_2}$ overlap different good sets, $B_{l_1} \cap B_{l_2} = \emptyset$. $\square$

**Lemma 7.** *Given a set of landmarks $L$ chosen by Landmark-Selection so that the condition in Lemma 3 is satisfied, there is some value of $r^* < 4d_{\text{crit}}$ such that each $X_i$ is contained in some ball $B_l$ around landmark $l \in L$ of radius $r^*$.*

*Proof.* For each good set $X_i$ choose a point $s_i \in X_i$ and a landmark $l_i \in L$ that satisfy $d(s_i, l_i) < 2d_{\text{crit}}$. The distance between $l_i$ and each point $x \in X_i$ satisfies $d(l_i, x) \leq d(l_i, s_i) + d(s_i, x) < 2d_{\text{crit}} + 2d_{\text{crit}} = 4d_{\text{crit}}$.

Consider $r^* = \max_{l_i} \max_{x \in X_i} d(l_i, x)$. Clearly, each $X_i$ is contained in a ball $B_{l_i}$ of radius $r^*$ and $r^* < 4d_{\text{crit}}$. $\square$

**Lemma 8.** *Suppose the distance between $c_i^*$ and any landmark $l$ in $C'_{\sigma(i)}$ satisfies $d(c_i^*, l) < 5d_{\text{crit}}$. Then given point $x \in C_i^*$ that satisfies $w_2(x) - w(x) \geq 17d_{\text{crit}}$, for any $l_1 \in C'_{\sigma(i)}$ and $l_2 \in C'_{\sigma(j \neq i)}$ it must be the case that $d(x, l_1) < d(x, l_2)$.*

*Proof.* We will show that $d(x, l_1) < w(x) + 5d_{\text{crit}}$ **(1)**, and $d(x, l_2) > w(x) + 12d_{\text{crit}}$ **(2)**. This implies that $d(x, l_1) < d(x, l_2)$.

To prove **(1)**, by the triangle inequality $d(x, l_1) \leq d(x, c_i^*) + d(c_i^*, l_1) = w(x) + d(c_i^*, l_1) < w(x) + 5d_{\text{crit}}$. To prove **(2)**, by the triangle inequality $d(x, c_j^*) \leq d(x, l_2) + d(l_2, c_j^*)$. It follows that $d(x, l_2) \geq d(x, c_j^*) - d(l_2, c_j^*)$. Since $d(x, c_j^*) \geq w_2(x)$ and $d(l_2, c_j^*) < 5d_{\text{crit}}$

we have
$$d(x, l_2) > w_2(x) - 5d_{\text{crit}}. \quad (1)$$

Moreover, since $w_2(x) - w(x) \geq 17d_{\text{crit}}$ we have
$$w_2(x) \geq 17d_{\text{crit}} + w(x). \quad (2)$$

Combining Equations 1 and 2 it follows that $d(x, l_2) > 17d_{cri} + w(x) - 5d_{\text{crit}} = w(x) + 12d_{\text{crit}}$. □

*Proof of Theorem 2.* Each cluster in the clustering $C' = \{C'_1, C'_2, \ldots C'_k\}$ output by *Expand-Landmarks* contains points from a distinct good set $X_i$. This clustering can exclude up to $b$ points, all of which may be good. Nonetheless, this means that $C'$ may disagree with $C^*$ on only the bad points and at most $b$ good points. The number of points that $C'$ and $C^*$ disagree on is therefore at most $2b = O(\epsilon n/\alpha)$. Thus, $C'$ is at least $O(\epsilon/\alpha)$-close to $C^*$, and at least $O(\epsilon/\alpha + \epsilon)$-close to $C_T$.

Moreover, $C'$ has an additional property that allows us to find a clustering that is $\epsilon$-close to $C_T$. If we use $\sigma$ to denote a bijection mapping each good set $X_i$ to the cluster $C'_{\sigma(i)}$ containing points from $X_i$, any landmark $l \in C'_{\sigma(i)}$ is closer than $5d_{\text{crit}}$ to $c^*_i$. We can use this observation to find all points that satisfy one of the properties of the good points: points $x$ such that $w_2(x) - w(x) \geq 17d_{\text{crit}}$. Let us call these points the *detectable* points. To clarify, the detectable points are those points that are much closer to their own cluster center than to any other cluster center in $C^*$, and the *good* points are a subset of the detectable points that are also very close to their own cluster center.

To find the detectable points using $C'$, we choose some landmark $l_i$ from each $C'_i$. For each point $x \in S$, we then insert $x$ into the cluster $C''_j$ for $j = \text{argmin}_i d(x, l_i)$. Lemma 8 argues that each detectable point in $C^*_i$ is closer to every landmark in $C'_{\sigma(i)}$ than to any landmark in $C'_{\sigma(j \neq i)}$. It follows that $C''$ and $C^*$ agree on all the detectable points. Since there are fewer than $(\epsilon - \epsilon^*)n$ points on which $C_T$ and $C^*$ agree that are not detectable, it follows that $\text{dist}(C'', C_T) < (\epsilon - \epsilon^*) + \text{dist}(C_T, C^*) = (\epsilon - \epsilon^*) + \epsilon^* = \epsilon$. □

## 4 Empirical Study

We use our *Landmark Clustering* algorithm to cluster proteins using sequence similarity. As mentioned in the Introduction, one versus all distance queries are particularly relevant in this setting because of sequence database search programs such as BLAST [1] (Basic Local Alignment Search Tool). BLAST aligns the queried sequence to sequences in the database, and produces a "bit score" for each alignment, which is a measure of its quality (we invert the bit score to make it a distance). However, BLAST does not consider alignments with some of the sequences in the database, in which case we assign distances of infinity to the corresponding sequences. We observe that if we define distances in this manner they almost form a metric in practice: when we draw random triplets of sequences and check the distances between them the triangle inequality is almost always satisfied. Moreover, BLAST is very successful at detecting sequence homology in large sequence databases, therefore it is plausible that clustering using these distances satisfies the $(c, \epsilon)$-property for some relevant clustering $C_T$.

We perform experiments on datasets obtained from two classification databases: Pfam [6] (version 24.0, October 2009) and SCOP [8] (version 1.75, June 2009). Both of these sources classify proteins by their evolutionary relatedness, therefore we can use their classifications as a ground truth to evaluate the clusterings produced by our algorithm and other methods.

Pfam classifies proteins using hidden Markov models (HMMs) that represent multiple sequence alignments. There are two levels in the Pfam classification hierarchy: family and clan. In our clustering experiments we compare with a classification at the family level because the relationships at the clan level are less likely to be discerned with sequence alignment. In each experiment we randomly select several large families (of size between 1000 and 10000) from Pfam-A (the manually curated part of the classification), retrieve the sequences of the proteins in these families, and use our *Landmark-Clustering* algorithm to cluster the dataset.

SCOP groups proteins on the basis of their 3D structures, so it only classifies proteins whose structure is known. Thus the datasets from SCOP are much smaller in size. The SCOP classification is also hierarchical: proteins are grouped by class, fold, superfamily, and family. We consider the classification at the superfamily level because this seems most appropriate given that we are only using sequence information. As with the Pfam data, in each experiment we create a dataset by randomly choosing several superfamilies (of size between 20 and 200), retrieve the sequences of the corresponding proteins, and use our *Landmark-Clustering* algorithm to cluster the dataset.

Once we cluster a particular dataset, we compare the clustering to the manual classification using the distance measure from the theoretical part of our work. To find the fraction of misclassified points under the optimal matching of clusters in $C$ to clusters in $C'$ we solve a minimum weight bipartite matching problem where the cost of matching $C_i$ to $C'_{\sigma(i)}$ is $|C_i - C'_{\sigma(i)}|/n$.

### 4.1 Choice of Parameters

To run *Landmark-Clustering*, we set $k$ using the number of clusters in the ground truth clustering. For each Pfam dataset we use $40k$ landmarks/queries, and for each SCOP dataset we use $30k$ landmarks/queries. In addition, our algorithm uses three parameters $(q, s_{\min}, n')$ whose value is set in the proof based on $\alpha$ and $\epsilon$, assuming that the clustering instance satisfies the $(1+\alpha, \epsilon)$-property. In practice we must choose some value for each parameter. In our experiments we set them as a function of the number of points in the dataset $(n)$, and the average size of the ground truth clusters $(\mu)$. We set $q = 2\mu$, $s_{\min} = 0.05\mu/0.1\mu$ for Pfam/SCOP datasets, and $n' = 0.5n$. Since the selection of landmarks is randomized, for each dataset we perform several clusterings, compare each to the ground truth, and report the median quality.

*Landmark-Clustering* is most sensitive to the $s_{\min}$ parameter, and will not report a clustering if $s_{\min}$ is too small or too large. We recommend trying several reasonable values of $s_{\min}$, in increasing or decreasing order, until you get a clustering and none of the clusters are too large. If you get a clustering where one of the clusters is very large, this likely means that several ground truth clusters have been merged. This may happen because $s_{\min}$ is too small causing balls of outliers to connect different cluster cores, or $s_{\min}$ is too large causing balls in different cluster cores to overlap.

The algorithm is less sensitive to the $n'$ parameter. However, if you set $n'$ too large some ground truth clusters may be merged, so we recommend using a smaller value ($0.5n \leq n' \leq 0.7n$) because all of the points are still clustered during the last step. Again, for some values of $n'$ the algorithm may not output a clustering, or output a clustering where some of the clusters are too large. Our algorithm is least sensitive to the $q$ parameter. Using more landmarks (if you can afford it) can make up for a poor choice of $q$.

### 4.2 Results

Figure 1 shows the results of our experiments on the Pfam datasets. One can see that for most of the datasets (other than datasets 7 and 9) we find a clustering that is almost identical to the ground truth. These datasets are very large, so as a benchmark for comparison we can only consider algorithms that use a comparable amount of distance information (since we do not have the full distance matrix). A natural choice is the following algorithm: randomly choose a set of landmarks $L$, $|L| = d$; embed each point in a $d$-dimensional space using distances to $L$; use $k$-means clustering in this space.

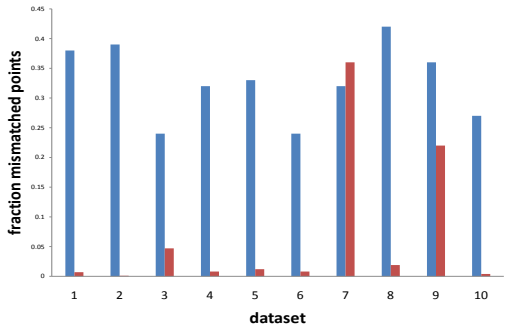

Figure 1: Comparing the performance of $k$-means in the embedded space (blue) and *Landmark-Clustering* (red) on 10 datasets from Pfam. Datasets **1-10** are created by randomly choosing 8 families from Pfam of size $s$, $1000 \leq s \leq 10000$.

Notice that this procedure uses exactly $d$ one versus all distance queries, so we can set $d$ equal to the number of queries used by our algorithm. We expect this procedure to work well, and indeed if you look at Figure 1 you can see that it finds reasonable clusterings. Still, the clusterings reported by this procedure do not match the Pfam classification exactly, showing that finding the exact Pfam partition is not trivial.

Figure 2 shows the results of our experiments on the SCOP datasets. These results are not as good, which is likely because the SCOP classification at the superfamily level is based on biochemical and structural evidence in addition to sequence evidence. By contrast, the Pfam classification is based entirely on sequence information. Still, because the SCOP datasets are much smaller, we can compare our algorithm to methods that require distances between all the points. In particular, Paccanaro et al. showed that spectral clustering using sequence data works well when applied to the proteins in SCOP [10]. Thus we use the exact method described in [10] as a benchmark for comparison on the SCOP datasets. Moreover, other than clustering randomly generated datasets from SCOP, we also consider the two main examples from [10], which are labeled **A** and **B** in the figure. From Figure 2 we can see that the performance of *Landmark-Clustering* is comparable to that of the spectral method, which is very good considering that the algorithm used by Paccanaro et al. significantly outperforms other clustering algorithms on this data [10]. Moreover, the spectral clustering algorithm requires the full distance matrix as input, and takes much longer to run.

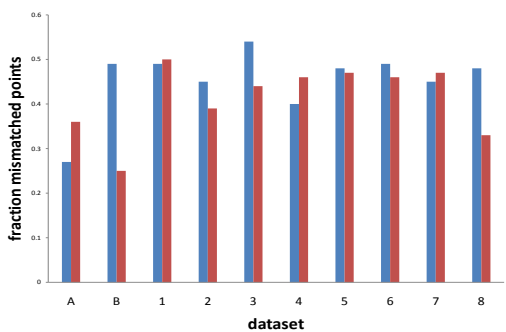

Figure 2: Comparing the performance of spectral clustering (blue) and *Landmark-Clustering* (red) on 10 datasets from SCOP. Datasets **A** and **B** are the two main examples from [10], the other datasets (**1-8**) are created by randomly choosing 8 superfamilies from SCOP of size $s$, $20 \leq s \leq 200$.

## 5 Conclusion and Open Questions

In this work we presented a new algorithm for clustering large datasets with limited distance information. As opposed to previous settings, our goal was not to approximate some objective function like the $k$-median objective, but to find clusterings close to the ground truth. We proved that our algorithm yields accurate clusterings with only a small number of one versus all distance queries, given a natural assumption about the structure of the clustering instance. This assumption has been previously analyzed in [3], but in the full distance information setting. By contrast, our algorithm uses only a small number of queries, it is much faster, and it has the same formal performance guarantees as the one introduced in [3].

To demonstrate the practical use of our algorithm, we clustered protein sequences using a sequence database search program as the one versus all query. We compared our results to gold standard manual classifications of protein evolutionary relatedness given in Pfam [6] and SCOP [8]. We find that our clusterings are comparable in accuracy to the classification given in Pfam. For SCOP our clusterings are as accurate as state of the art methods, which take longer to run and require the full distance matrix as input.

Our main theoretical guarantee assumes large target clusters. It would be interesting to design a provably correct algorithm for the case of small clusters as well.

### Acknowledgements

The authors would like to thank the reviewers for constructive feedback that greatly helped to improve the implementation of the algorithm.

Konstantin Voevodski was supported by an IGERT Fellowship through NSF grant DGE-0221680 awarded to the ACES Training Program at BU Center for Computational Science. Maria Florina Balcan was supported in part by NSF grant CCF-0953192, by ONR grant N00014-09-1-0751 and AFOSR grant FA9550-09-1-0538. Heiko Röglin was supported by a Veni grant from the Netherlands Organisation for Scientific Research. Shang-Hua Teng was supported in part by NSF grant CCR-0635102.